\PassOptionsToPackage{table,xcdraw,dvipsnames}{xcolor}

\documentclass[]{fairmeta}
\usepackage{tcolorbox}
\usepackage{graphicx}	
\usepackage{amsmath}	
\usepackage{amssymb}	
\usepackage{booktabs}
\usepackage{microtype}
\usepackage{epsfig}
\usepackage[table,xcdraw,dvipsnames]{xcolor}
\usepackage{caption}
\usepackage{float}
\usepackage{placeins}
\usepackage{color, colortbl}
\usepackage{stfloats}
\usepackage{enumitem}
\usepackage{tabularx}
\usepackage{xstring}
\usepackage{multirow}
\usepackage{xspace}
\usepackage{url}
\usepackage{subcaption}
\usepackage{xcolor}
\usepackage[hang,flushmargin]{footmisc}
\usepackage{amsmath}
\usepackage{graphicx}
% ...

\newcommand{\R}[1]{{%
    \textbf{%
        \ifstrequal{#1}{1}{\textcolor{red}{R#1}}{%
        \ifstrequal{#1}{2}{\textcolor{blue}{R#1}}{%
        \ifstrequal{#1}{3}{\textcolor{magenta}{R#1}}{%
        \ifstrequal{#1}{4}{\textcolor{teal}{R#1}}{%
                           \textcolor{cyan}{R#1}%
        }}}}%
    }%
}}

\newtcbox{\hlprimarytab}{on line, box align=base, colback=BlueGreen!20, size=fbox, before upper=\strut, top=-1.5pt, bottom=-1.5pt, left=-1pt, right=-1pt, boxrule=0pt}

\newtcbox{\hlsecondarytab}{on line, box align=base, colback=WildStrawberry!10,size=fbox, before upper=\strut, top=-1.5pt, bottom=-1.5pt, left=-2pt, right=-2pt, boxrule=0pt}

\newcommand{\orange}[1]{{\hlsecondarytab{#1}}}
\newcommand{\blue}[1]{{\hlprimarytab{#1}}}  % Add packages to _macros.tex
\newcommand{\name}{{LMFusion}\xspace}

\title{\name: Adapting Pretrained Language Models for Multimodal Generation}

\author[1,*]{Weijia Shi}
\author[1,*]{Xiaochuang Han } 
\author[]{Chunting Zhou} 
\author[3]{Weixin Liang}
\author[2]{Xi Victoria Lin}
\author[1,2]{Luke Zettlemoyer}
\author[2]{Lili Yu}
\affiliation[1]{University of Washington}
\affiliation[2]{FAIR at Meta}
\affiliation[3]{Stanford University}

\contribution[*]{Joint first author. Order randomly determined. Work done while at Meta.}
\abstract{We present \name , a framework for empowering pretrained text-only large language models (LLMs) with multimodal generative capabilities, enabling them to understand and generate both text and images in arbitrary sequences. 
\name leverages existing Llama-3's weights for processing texts autoregressively while introducing additional and parallel transformer modules for processing images with diffusion. 
During training, the data from each modality is routed to its dedicated modules: modality-specific feedforward layers, query-key-value projections, and normalization layers process each modality independently, while the shared self-attention layers allow interactions across text and image features. 
By freezing the text-specific modules and only training the image-specific modules, \name preserves the language capabilities of text-only LLMs while developing strong visual understanding and generation abilities. Compared to methods that pretrain multimodal generative models from scratch, our experiments demonstrate that, \name improves image understanding by 20\% and image generation by 3.6\% using only 50\% of the FLOPs while maintaining Llama-3’s language capabilities. 
We also demonstrate that this framework can adapt existing vision-language models with multimodal generation ability. 
Overall, this framework not only leverages existing computational investments in text-only LLMs but also enables the parallel development of language and vision capabilities, presenting a promising direction for efficient multimodal model development. }

\correspondence{Weijia Shi \email{swj0419@uw.edu}, Xiaochuang Han \email{xhan77@uw.edu}, Lili Yu \email{liliyu@meta.com}}
\date{\today}

\begin{document}
\maketitle

\begin{figure*}[ht]
    \centering
    \includegraphics[width=\linewidth]{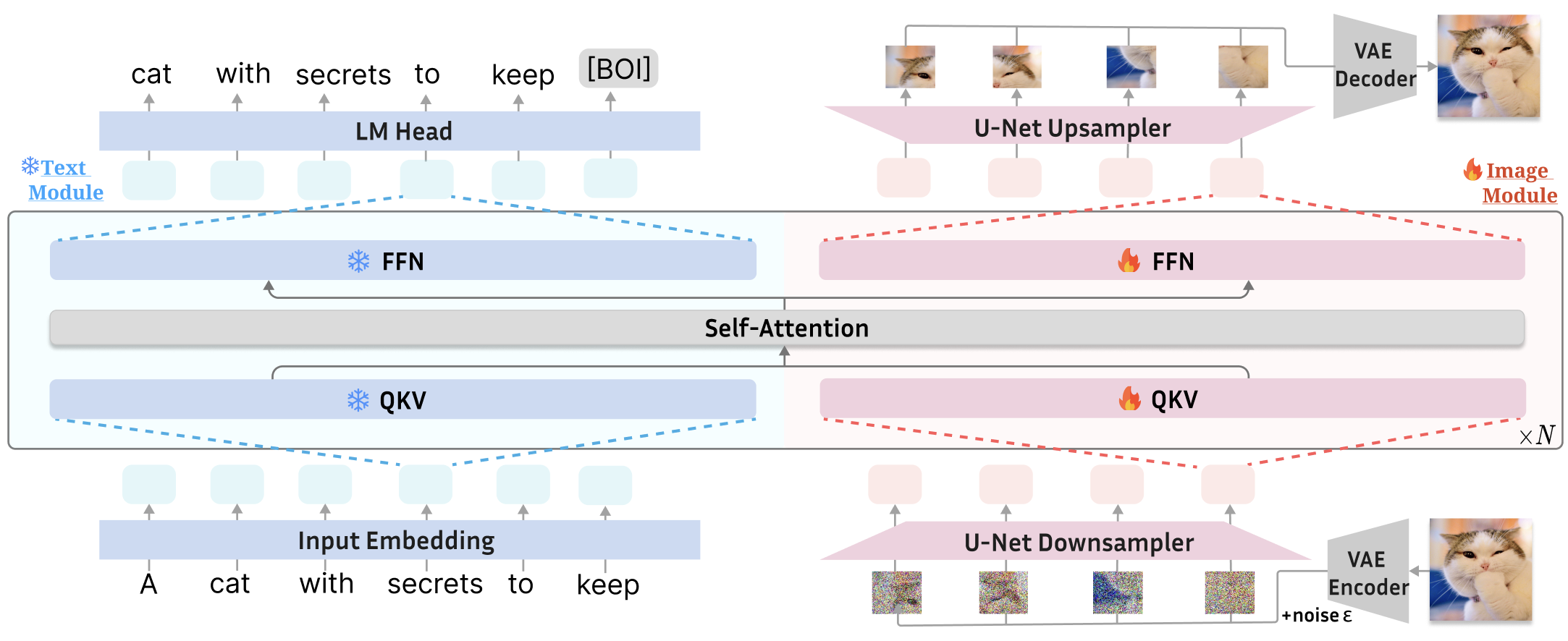}
    \caption{\textbf{Overview of \name}. It uses modality-specific FFNs and QKV projections to process text and image data separately: the text ``A cat with secrets to keep'' goes to the \blue{text module}, while the image patches of the cat goes to the \orange{image module}. In the self-attention layer, text and image representations can attend to all previous contexts across the modality boundaries. Both modules are initialized from Llama-3, with the text module frozen to preserve language capabilities while the image module trained on image data. Layer normalization and residual connections are folded into the QKV and FFN modules. A special \texttt{BOI} token separates different modalities in the sequence.  
    }
    \label{fig:teaser}
    \vspace{-0.5em}
\end{figure*}

% \begin{figure*}[ht]
%     \centering
%     \includegraphics[width=\linewidth]{figs/teaser.png}
%     \caption{\textbf{Overview of \name}. It uses modality-specific FFNs and QKV projections to process text and image data separately: the text ``A cat with secrets to keep'' goes to the \blue{text module}, while the image patches of the cat goes to the \orange{image module}. In the self-attention layer, text and image representations can attend to all previous contexts across the modality boundaries. Both modules are initialized from Llama-3, with the text module frozen to preserve language capabilities while the image module trained on image data. Layer normalization and residual connections are folded into the QKV and FFN modules. A special \texttt{BOI} token separates different modalities in the sequence.  
%     }
%     \label{fig:teaser}
%     \vspace{-0.5em}
% \end{figure*}

\begin{figure*}[t]
    \centering
    \includegraphics[width=\linewidth]{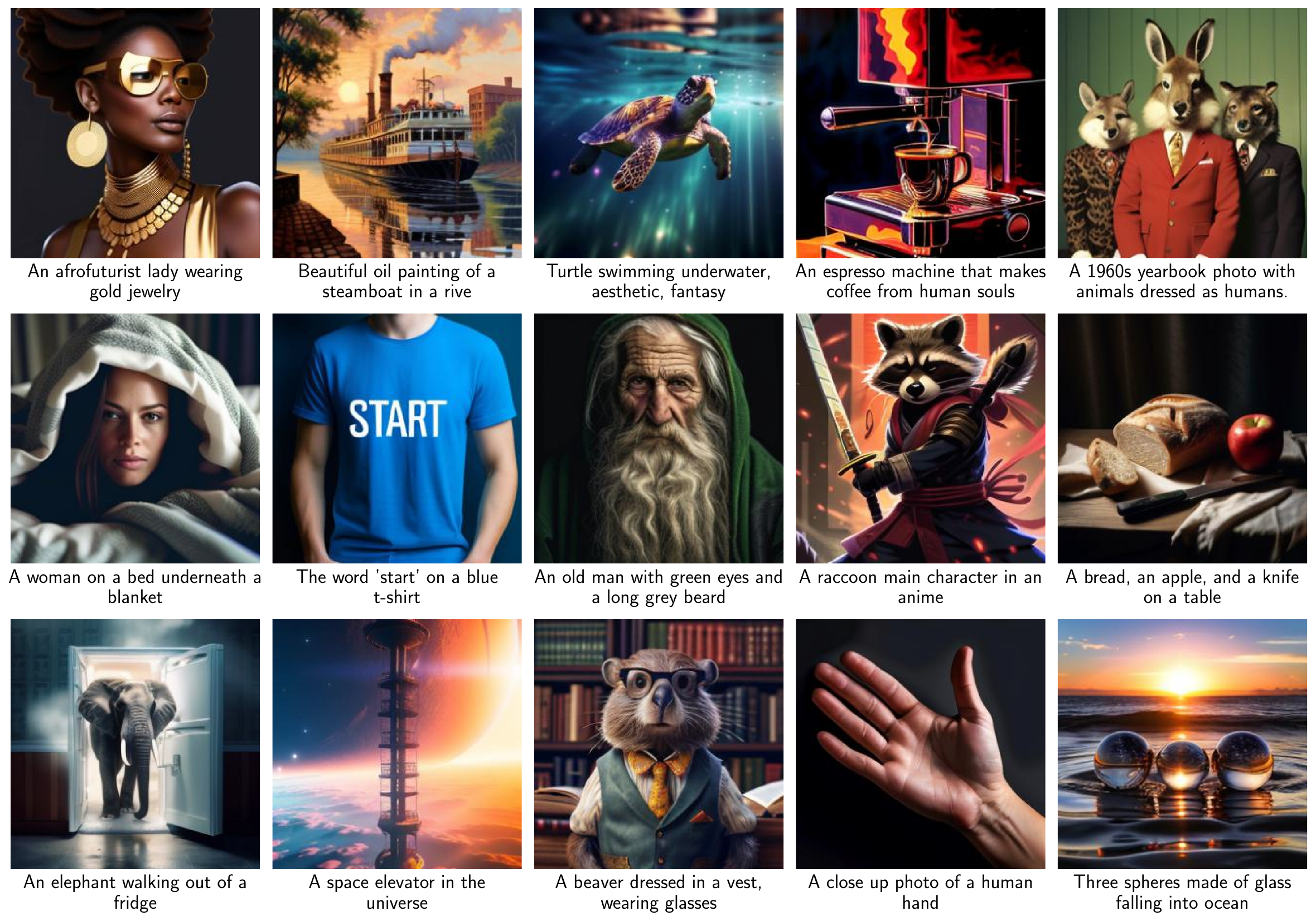}
    \caption{Generated images from \name fine-tuned on aesthetically appealing images for improved quality. 
    }
    \label{fig:final_pick}
    \vspace{-1.0em}
\end{figure*}

\section{Introduction} 
Over the past few years, we have seen significant progress in multimodal generative models capable of understanding and generating interleaved text and images in arbitrary sequences \citep{dong2023dreamllm, koh2024generating, lin2024momaefficientearlyfusionpretraining}. 
Models like Transfusion \citep{transfusion}, Chameleon \citep{team2024chameleon}, and Unified-IO \citep{lu2022unified, lu2024unified} demonstrate the potential of unified architectures that seamlessly handle both image and text modalities. 
However, these models typically train from scratch, demanding significant computational resources to achieve proficiency across all modalities. 
The computational cost of mastering even a single modality is substantial---training a state-of-the-art text-only large language models (LLMs) like Llama-3 \citep{llama3} requires training over 15 trillion tokens.

% Given these computational demands, we investigate an alternative paradigm: reusing and adapting existing pretrained LLMs for multimodal understanding and generation. 
% This approach raises a fundamental research question: 
Given these computational demands, we investigate an alternative paradigm that reuses and adapts existing pretrained LLMs \citep{ge2023making,sun2023generative,wu2024vila}. 
We address a fundamental research question: 
\emph{How to preserve the text-only performance of pretrained LLMs while equipping them with visual understanding and generation abilities?}
% \swj{1. visual understanding and generation or multimodal understanding and generation? 2. reptitive}
% \han{Given these computational demands, we investigate an alternative paradigm that reuses and adapts existing pretrained LLMs. We address a fundamental research question: \emph{How to preserve the text-only performance of pretrained LLMs while equipping them with visual understanding and generation abilities?}}  
% \lili{I like HX's much better, let's not call it a "new" paradigm. We should also cite paper like, emu2, seedx, janusflow} 
% \han{I added a bunch of references that reuse any LLMs} \swj{can this preserve LM performance? }
Our experiments show that naive finetuning of pretrained text-only LLMs on multimodal data leads to significant degradation of their language processing capabilities.

To address this challenge, we introduce \textbf{\name}, a framework that enhances a pretrained text-only LLM, Llama-3 \citep{llama3} with multimodal capabilities by building upon the recipe of Transfusion \citep{transfusion}. 
Drawing from recent and parallel work on modality separation \citep{shen2023scaling, Chen2023EVEEV, liang2024mixtureoftransformers, liu2024playgroundv3improvingtexttoimage},
\name integrates the original Llama modules pretrained for language processing while introducing additional dedicated transformer modules for visual understanding and generation tasks. 
As shown in \autoref{fig:teaser}, we employ modality-specific query-key-value (QKV) projections and feed-forward networks (FFNs) to process text and image data separately while still allowing for cross-modal interactions in the joint self-attention layer. %\swj{this part maye not be clear, @han}
By freezing the text modules while finetuning the image modules, we preserve the language-only capabilities of pretrained LLMs while giving a head start to the learning of visual understanding and generation. 
Compared to pretraining multimodal generative models from scratch, this approach avoids the need to include text-only data in the training process, significantly reducing the computational demands.

% This approach significantly reduces computational demands by leveraging existing pretrained LLMs' capabilities without the need to include text-only pretraining data in the training process.

% , with a seminal multimodal training recipe, Transfusion \citep{transfusion}. 

To evaluate the effectiveness of our approach, we conduct comprehensive experiments comparing \name with Transfusion in controlled settings. Specifically, we initialize our \name architecture with a pretrained Llama-3 8B model \citep{llama3} and continue training on the same image data 
% \han{changed; original: multimodal data} 
% \lili{I think the freezing text model counts as the final LlaMaFusion model, then it is not the same data, we only use image data} 
as in Transfusion \citep{transfusion}. 
% \name demonstrates strong image generation and understanding capabilities while maintaining the strong text-only task performance of the original Llama-3 model. 
Compared to Transfusion, \name achieves a 20\% improvement in image understanding, 3.6\% improvement in image generation while using only 50\% of the FLOPs. 
% with only 50\% of the FLOPs, and 
It also preserves Llama-3's text-only performance that outperforms Transfusion by 11.6\%. \autoref{fig:final_pick} presents images generated by \name. Additionally, we further demonstrate that this framework can adapt existing vision-language models (e.g., LLaVA) with multimodal generation ability.

Through ablation studies, we analyze the key architectural decision for \name: 
separating both self-attention and FFNs for different modality data while freezing weights for the pretrained language modality. 
% architectural decisions that influence \name’s performance. 
% We find one critical design choice is deep modality separation -- \textit{separating both self-attention and FFNs} for different modality data. 
We show that naive finetuning of the dense pretrained LLMs on multimodal data (\textit{no separation}) leads to a catastrophic forgetting of their original language capabilities.  Furthermore, deep separation proves to be more effective than shallow separation (\textit{using modality-specific FFNs only}), with both approaches outperforming models with no separation. 
% \swj{This concept of modality separation } have been explore d parallel work (cite mot) investigates the separation of model components for multiple modalities. In this work, we share a same basic intuition but show its power/suitability in continuing training LLMs for image diffusion capabilities.

% We find one critical design choice is the use of deep modality separation -- separating both self-attention and FFNs: 
% Notably, naive finetuning of the dense pretrained LLMs on multimodal data (No separation) leads to a catastrophic forgetting of their original language capabilities. Deep separation is superior to shallow separation with dedicated FFNs only, with both outperforming no sepration. 
% (despite mitigations like modality-specific learning rates). \swj{not sure if this part is clear to the reader}
% We find that a deep modality weight separation---with both modality-specific self-attention and FFNs---is superior to shallow separation with dedicated FFNs only, with both outperforming a vanilla dense model without modality separation. %\swj{this part may not be clear or repetitive}

Overall, \name has the following key features: (1) \textit{Compute reuse}:  It leverages existing computational investments in text-only LLMs when developing multimodal generative models.
This eliminates the need to retrain on text-only data, significantly reducing computational demands. 
(2) \textit{Performance preservation and transfer}: It completely preserves the strong text-only performance of pretrained LLMs and facilitates a better learning of image understanding and generation in the multimodal generative setup.

\section{Background: Transfusion}
\label{sec:background}

Transfusion \citep{transfusion} is a single unified multimodal model that is capable of text generation, image understanding, and image generation tasks, by jointly predicting next tokens in language and diffusing image representations. 
Given a multimodal input $(\boldsymbol{x}^\textit{txt}, \boldsymbol{x}^\textit{img})$, the Transfusion model jointly learns to do \textit{language modeling} (\S \ref{sec:transfusion-language}) on $\boldsymbol{x}^\textit{txt}$ and \textit{image diffusion} (\S \ref{sec:transfusion-image}) on $\boldsymbol{x}^\textit{img}$.
 Its architecture is same as a standard Transformer \citep{vaswani2017attention} with an additional U-Net structure \citep{ronneberger2015u} that projects image representations down and up before and after diffusion.

\subsection{Language Modeling} \label{sec:transfusion-language}
% \swj{add a footnote that vector is bold? }
Given a sequence of discrete language tokens $\boldsymbol{x}^\textit{txt} = x_{1}^\textit{txt}, \ldots , x_{N}^\textit{txt}$, a language model $\theta$ represents its joint probability by $P(\boldsymbol{x}^\textit{txt}) = \prod_{i=1}^N P_\theta(x_i^\textit{txt} \mid \boldsymbol{x}_{<i}^\textit{txt})$. 
This formulation sets up an autoregressive task, where each token $x_i^\textit{txt}$ is predicted based on its preceding tokens $\boldsymbol{x}_{<i}^\textit{txt}$. The language model is learned by minimizing the cross-entropy between $P_\theta$ and the observed data distribution, which is commonly referred to as the LM loss: 
\begin{align} \label{eq: lm_loss}
\mathcal{L}_\text{LM} = \mathbb{E}_{x_{i}^\textit{txt}}[-\log P_{\theta}(x_{i}^\textit{txt} \mid \boldsymbol{x}_{<i}^\textit{txt}, \boldsymbol{x}^\textit{img})]
\end{align}

Optionally, if there exists image data preceding the language tokens (e.g., image-caption data), Transfusion adds the representation of $\boldsymbol{x}^\textit{img}$ as additional condition to the objective. More details of representing $\boldsymbol{x}^\textit{img}$ are presented below.

% \subsection{Model arthtecture}

% {\color{blue} [Han: if we need to talk about exact model architecture here, maybe move the Transfusion paragraph in Sec. 4 here?]}

\subsection{Image Diffusion} \label{sec:transfusion-image}
% Given a multimodal input sequence consisting of a caption $x_\textit{txt}$ and image  $x_\textit{image}$.
% % $(x_\textit{txt}, x_\textit{image})$
% % = (x_1^\textit{txt}, \ldots, x_N^\textit{txt}, x_1^\textit{img}, \ldots, x_M^\textit{img})$,
% , Tranfusion optimizes two objectives: autoregressive \textit{language modeling} on $\boldsymbol{x}^\textit{txt}$ and \textit{image diffusion} on $\boldsymbol{x}^\textit{img}$. 
Given a raw image, 
Transfusion first encodes the image into a sequence of continuous latent representation $\boldsymbol{x}^\textit{img}$ with a pretrained and frozen VAE tokenizer \citep{kingma2013auto}. 
It then employs Denoising Diffusion Probabilistic Models (i.e., DDPM) to learn to reverse a gradual noise-addition process added in the forward process \citep{ho2020denoising}. 
% {\color{red} It further reduced the image representaiton dimension using U-Net up and down sampler. (Figure XX).} 
% For image diffusion, Transfusion models a sequence of continuous latent embeddings of images $\boldsymbol{x}^\textit{img}$, encoded by a pretrained and frozen VAE tokenizer. 
In the forward diffusion process, a Gaussian noise $\boldsymbol{\epsilon} \sim \mathcal{N}(\boldsymbol{0}, \mathbf{I})$ is added to the image representation $\boldsymbol{x}^\textit{img}$ over $T$ steps, creating a sequence of noisy image representations $\boldsymbol{x}_0, \boldsymbol{x}_1, ..., \boldsymbol{x}_T$. Specifically, at each step $t$, the noisy image representation is given by:
\begin{align}
\boldsymbol{x}_t^{\textit{img}} = \sqrt{\bar{\alpha}_t}\boldsymbol{x}^\textit{img} + \sqrt{1-\bar{\alpha}_t} \boldsymbol{\epsilon}
\end{align}
Here $\bar{\alpha}_t$ follows a common cosine schedule \citep{nichol2021improved}. 

% For the diffusion process, a Gaussian noise $\boldsymbol{\epsilon} \sim \mathcal{N}(\boldsymbol{0}, \mathbf{I})$ is added to the image representation $\boldsymbol{x}^\textit{img}$, and 
In the reverse process, the diffusion model $\boldsymbol{\epsilon}_{\theta}(\cdot)$ with parameters $\theta$ learns to predict the added noise $\boldsymbol{\epsilon}$ given the noisy data $\boldsymbol{x}_t^{\textit{img}}$ at timestep $t$ and a context $\boldsymbol{x}^\textit{txt}$ that can include text prompts such as captions to the image diffusion: \footnote{Similar to $\boldsymbol{x}^\textit{txt}$, this context can also include image representations $\boldsymbol{x}^\textit{img}$ under an image editing setup. We omit it in the notation for simplicity.} 
% via DDPM objective
% $\boldsymbol{\epsilon}_{\theta}(\cdot)$ 
% This process is conditioned on a uniformly sampled time step $t$ which determines the level of noise added to the input, and a context $\boldsymbol{x}^\textit{txt}$ that can include textual prompts to the image diffusion.\footnote{Similar to $\boldsymbol{x}^\textit{txt}$, this context can also include image representations $\boldsymbol{x}^\textit{img}$ under an image editing setup. We omit it in the notation for simplicity.}  
% Transfusion learns a reverse diffusion via the DDPM objective that predicts the added noise. This process is conditioned on a uniformly sampled time step $t$ which determines the level of noise added to the input, and a context $\boldsymbol{x}^\textit{txt}$ that can include textual prompts to the image diffusion.\footnote{Similar to $\boldsymbol{x}^\textit{txt}$, this context can also include image representations $\boldsymbol{x}^\textit{img}$ under an image editing setup. We omit it in the notation for simplicity.} 
\begin{align} \label{eq: ddpm_loss}
    \mathcal{L}_\text{DDPM} = \mathbb{E}_{\boldsymbol{x}^\textit{img}, t, \boldsymbol{\epsilon}} [\lVert \boldsymbol{\epsilon} - \boldsymbol{\epsilon}_{\theta}(\boldsymbol{x}_t^{\textit{img}}, t, \boldsymbol{x}^\textit{txt}) \rVert_2^2]
\end{align}
% We use $\boldsymbol{\epsilon}_{\theta}(\cdot)$ to denote the model that predicts the added noise $\boldsymbol{\epsilon}$. 
% \swj{should we refer to figure 1?}
% As shown in \autoref{fig:teaser}, 
The Transfusion architecture contains U-Net downsampler and upsampler to reduce the dimension of $\boldsymbol{x}^{\textit{img}}$. The U-Net downsampler transforms the image into fewer patches before the main Transformer modules while the upsampler projects them back to the original dimension of $\boldsymbol{x}^{\textit{img}}$ after the Transformer. 

% In transfusion, the model architecture and parameters $\theta$ are instantiated through a U-Net downsampler that reduces the dimension of $\boldsymbol{x}^\textit{img}$, a main Transformer that is shared with the language modeling objective, and a U-Net upsampler that projects back to the original dimension of $\boldsymbol{x}^\textit{img}$. 

% Looking inside the model parameters $\theta$, the main Transformer blocks and the U-Net downsampler are optimized under both the LM loss and the diffusion loss, whereas the U-Net upsampler receives gradients only from the diffusion loss. 

\subsection{Training Objective}
During training, Transfusion is optimized to predict both the LM loss on the text input $\boldsymbol{x}^\textit{txt}$ and the diffusion loss on the image input $\boldsymbol{x}^\textit{img}$. These two losses are combined using a hyperparameter $\lambda$:
\begin{align} \label{eq: transfusion}
\mathcal{L}_\text{Transfusion} = \mathcal{L}_\text{LM} + \lambda \cdot \mathcal{L}_\text{DDPM}
\end{align}

\section{\name}
\label{sec:method}
One notable feature of Transfusion is that it has the same architecture as mainstream LLMs (e.g., Llama \citep{touvron2023llama}) while being capable of text generation, image understanding, and image generation together, through an end-to-end training (\autoref{eq: transfusion}). 
% after end-to-end training with the objective in \autoref{eq: transfusion}. 
\citet{transfusion} trains Transfusion from scratch using language-only and image-caption data. 
However, such training from scratch requires substantial computational resources, and its performance on language-only tasks still lags behind the pretrained, text-only LLMs. 

% In this work, we aim to explore approaches for effectively re-using pretrained, language-only LLMs to make them able to understand and generate images.
% Specifically, we build on the open-weight LLM, LLaMA-3 \citep{llama3}, by making them able to 
% The key challenge is to preserve the text-only performance of Llama-3 when optimizing for the image understanding and image generation abilities. 

In this work, we aim to effectively adapt pretrained, text-only LLMs to handle image understanding and generation tasks. 
Specifically, we build on an open-weight LLM, Llama-3 \citep{llama3}, and continue training it with the Transfusion objectives to handle both modalities. 
% Specifically, we build on an open-weight LLM, LLaMA-3 \citep{llama3}, extending its capabilities to seamlessly generate both discrete text and continuous image modalities. 
Since Transfusion uses shared parameters for its language modeling and image diffusion objectives, the key challenge is to prevent Llama-3’s strong text-only performance from dropping while optimizing for its new image capabilities.

\subsection{Model Architecture}
In response to the challenge above, we propose \name, a framework that combines a pretrained, text-only Llama model with a dedicated image transformer for visual generation and understanding, enabling each modality to be processed through independent weights. By freezing the text modules while finetuning the visual modules, we preserve its language-only capabilities while giving the learning of visual understanding and generation a boost start. 
% can leverage the strong language capabilities of the pretrained LLM while endowing it with exceptional visual understanding and generation abilities.

% We now propose and define each component in LlamaFusion. 
\name is a decoder-only model consisting of $N$ transformer layers. As shown in \autoref{fig:teaser}, central to the design are the modality-specific attention layer and Feed-Forward Network (FFN), each handling only data from its corresponding modality. 
Without loss of generality, we describe \name below in a configuration with a single transformer layer, folding residual connections and layer normalization directly into the self-attention and FFN. 
The inputs to the model are text tokens $\boldsymbol{x}^\textit{txt}$ and noisy image representations $\boldsymbol{x}_t^{\textit{img}} = \sqrt{\bar{\alpha}_t}\boldsymbol{x}^\textit{img} + \sqrt{1-\bar{\alpha}_t} \boldsymbol{\epsilon}$. 
% and the current timestep $t$. 
% Below we instantiate the model parameters $\theta$ in the probability estimator $P_{\theta}(\boldsymbol{x}^\textit{txt}; \boldsymbol{x}^\textit{img})$ and noise predictor $\boldsymbol{\epsilon}_{\theta}(\sqrt{\bar{\alpha}_t}\boldsymbol{x}^\textit{img} + \sqrt{1-\bar{\alpha}_t} \boldsymbol{\epsilon}, t, \boldsymbol{x}^\textit{txt})$, as described in \autoref{sec:background}. 
We use {\blue{blue}} for text-specific modules and {\orange{red}} for image-specific modules. 

\paragraph{Input projection}
The input text tokens $\boldsymbol{x}^\textit{txt}$ are projected by a linear embedding layer to a sequence of text hidden states $\boldsymbol{h}_\text{in}^\textit{txt}$. The noisy image $\boldsymbol{x}_t^{\textit{img}}$ are projected to a sequence of image representations $\boldsymbol{h}_\text{in}^\textit{img}$ via a U-Net downsampler. 
\begin{align}
    \boldsymbol{h}_\text{in}^\textit{txt} =& \blue{$\operatorname{Proj}_{\text{text}}$}(\boldsymbol{x}^\textit{txt}) \\ 
    \boldsymbol{h}_\text{in}^\textit{img} =& \orange{$\operatorname{UNet-Down}_{\text{img}}$}(\boldsymbol{x}_t^{\text{img}}, t)
\end{align}
Then the text hidden states $\boldsymbol{h}_\text{in}^\textit{txt}$ or image hidden states $\boldsymbol{h}_\text{in}^\textit{img}$ are fed into the following attention layer. 

\paragraph{Modality-specific self-attention} 
We create separate attention matrices for each modality.
% the modality-specific attention layer with separate $Q, K, V$ matrices for image and text modality. 
% Specifically, the text hidden states $\boldsymbol{h}_\text{in}^\textit{txt}$ and image hidden states $\boldsymbol{h}_\text{in}^\textit{img}$ are converted to text queries, keys, values, and image queries, keys, values, via separate $Q, K, V$ matrices. The pre-attention layer normalization is also modality-specific and folded into $\operatorname{QKV}$ functions. 
Specifically, the text hidden states $\boldsymbol{h}_\text{in}^\textit{txt}$ and image hidden states $\boldsymbol{h}_\text{in}^\textit{img}$ are converted into their respective queries, keys, and values via separate $Q, K, V$ matrices. The pre-attention layer normalization is also modality-specific and is folded into the $\operatorname{QKV}$ functions. 
\begin{align}
    \boldsymbol{h}_\text{Q}^\textit{txt}, \boldsymbol{h}_\text{K}^\textit{txt}, \boldsymbol{h}_\text{V}^\textit{txt}=& \blue{$\operatorname{QKV}_\text{text}$}(\boldsymbol{h}_\text{in}^\textit{txt}) \\
    \boldsymbol{h}_\text{Q}^\textit{img}, \boldsymbol{h}_\text{K}^\textit{img}, \boldsymbol{h}_\text{V}^\textit{img} =& \orange{$\operatorname{QKV}_\text{img}$}(\boldsymbol{h}_\text{in}^\textit{img})
\end{align}
\noindent
We enable cross-modal attention by concatenating the queries, keys, and values from both image and text modalities into unified sequences. The attention-weighted values at text and image token positions are then projected back into the hidden state dimension using separate $\operatorname{O}$ weights for each modality.
% The attention-weighted values at positions of text and image tokens are then projected back to the hidden states dimension, again via separate weights. 
% The queries, keys, values for texts, and queries, keys, values for images are then concatenated to unified query, key, and value sequences. 
% This subsequently enables self-attention both within and across the modalities. 
\begin{align}
    \boldsymbol{h}_\text{O}^\textit{txt} =& \blue{$\operatorname{O}_{\text{text}}$}(\operatorname{softmax}(\frac{\boldsymbol{h}_\text{Q}^\textit{txt} [\boldsymbol{h}_\text{K}^\textit{img} \circ \boldsymbol{h}_\text{K}^\textit{txt}]^T + M}{\sqrt{d}}) [\boldsymbol{h}_\text{V}^\textit{img} \circ \boldsymbol{h}_\text{V}^\textit{txt}]) \\
    \boldsymbol{h}_\text{O}^\textit{img} =& \orange{$\operatorname{O}_{\text{img}}$}(\operatorname{softmax}(\frac{\boldsymbol{h}_\text{Q}^\textit{img} [\boldsymbol{h}_\text{K}^\textit{txt} \circ \boldsymbol{h}_\text{K}^\textit{img}]^T + M}{\sqrt{d}}) [\boldsymbol{h}_\text{V}^\textit{txt} \circ \boldsymbol{h}_\text{V}^\textit{img}])
\end{align}
where $\circ$ denotes concatenation. $M$ represents a hybrid attention mask same as in Transfusion \citep{transfusion} with a causal mask applied to text tokens and a bi-directional mask applied to image tokens. 
This design allows for self-attention within and across modalities, encouraging cross-modality integrations. 

% \begin{align}
%     \boldsymbol{h}_\text{O}^\textit{txt} =& \blue{$\operatorname{Attn-O}_{\text{text}}$}(\boldsymbol{h}_\text{Q, K, V}^\textit{txt}; \boldsymbol{h}_\text{K, V}^\textit{img}) \\
%     \boldsymbol{h}_\text{O}^\textit{img} =& \orange{$\operatorname{Attn-O}_{\text{img}}$}(\boldsymbol{h}_\text{Q, K, V}^\textit{img}; \boldsymbol{h}_\text{K, V}^\textit{txt})
% \end{align}

\paragraph{Modality-specific feed-forward network}
After the attention layer, we employ modality-specific FFNs to process text and image data separately. The pre-FFN layer normalization is also modality-specific and is folded in the $\operatorname{FFN}$ functions. 
% The text and image hidden states after self-attention are transformed with feed-forward networks corresponding to the modality. 
\begin{align}
    \boldsymbol{h}_\text{FFN}^\textit{txt} =& \blue{$\operatorname{FFN}_{\text{text}}$}(\boldsymbol{h}_\text{O}^\textit{txt}) \\
    \boldsymbol{h}_\text{FFN}^\textit{img} =& \orange{$\operatorname{FFN}_{\text{img}}$}(\boldsymbol{h}_\text{O}^\textit{img})
\end{align}

\paragraph{Output projection}
Finally, after $N$ layers of self-attention and FFNs, the resulting hidden states are projected either to logits in text via language model's output layer, or to predicted noise in image via a U-Net upsampler. 
\begin{align}
    \boldsymbol{p}_\text{logits} =& \blue{$\operatorname{LM-Head}_{\text{text}}$}(\boldsymbol{h}_\text{FFN}^\textit{txt}) \\
    \boldsymbol{\epsilon}_\text{pred} =& \orange{$\operatorname{UNet-Up}_{\text{img}}$}(\boldsymbol{h}_\text{FFN}^\textit{img}, t, \boldsymbol{h}_\text{in}^\textit{img})
\end{align}

\medskip

Same as Transfusion, the output $\boldsymbol{p}_\text{logits}$ and $\boldsymbol{\epsilon}_\text{pred}$ are passed through the language modeling loss (\autoref{eq: lm_loss}) and DDPM loss (\autoref{eq: ddpm_loss}) respectively. 
All parameters in the text modules along with self-attention and FFN parameters in the image modules are initialized from the pretrained Llama model. 
 During optimization, we \textbf{\textit{decouple the learning rates}} for the text and image parameter groups: a text learning rate, $\eta_{\text{text}}$, is used for \{\blue{\small 
$\operatorname{Proj}_{\text{text}}$,
$\operatorname{QKV}_\text{text}$,
$\operatorname{O}_{\text{text}}$,
$\operatorname{FFN}_{\text{text}}$,
$\operatorname{LM-Head}_{\text{text}} $
}\}
, and an image learning rate, $\eta_{\text{img}}$, for 
\{\orange{\small
$ \operatorname{UNet-Down}_{\text{img}}$,
$\operatorname{QKV}_\text{img}$,
$\operatorname{O}_{\text{img}}$,
$\operatorname{FFN}_{\text{img}}$,
$\operatorname{UNet-Up}_{\text{img}}$}\}
.
To preserve the model's performance on text-only benchmarks, we use $\eta_{\text{text}}=0$ (freezing text modules) for our main experiments and explore different configurations in \S \ref{sec:analysis}. 

\section{Experiments}
In this section, we describe the details of our training setup (\S \ref{sec: trainsetup}) and evaluation setup (\S \ref{sec: evalsetup}). 
Results in \S \ref{sec: results} show that \name outperforms Transfusion trained from scratch in the FLOPs match setting on text-only, image understanding and generation benchmarks. 
 % \swj{not sure compute match is the right word} \lili{Flops match?}

 % In this section, we first describe the details of our training setup  (\S\ref{sec:trainsetup}) and our evaluation setup (\S\ref{sec:evalsetup}). The results in \S\ref{sec:results} demonstrate that \name outperforms Transfusion trained from scratch in the FLOPs-matched setting across text-only, image understanding, and generation benchmarks.

\subsection{Training Setup} \label{sec: trainsetup}
\paragraph{Data}
% Comparable with Transfusion \citep{transfusion}, we use the same collection of 380M Shutterstock image-caption data, where each image is center-cropped and resized to $256 \times 256$ pixels. Following Transfusion, we order the captions before images (i.e., emphasizing image generation conditioned on texts) 80\% of the time, and order the images before captions for the rest. 

Following Transfusion \citep{transfusion}, we use the same collection of 380M Shutterstock image-caption data, where each image is center-cropped and resized to $256 \times 256$ pixels. We order the captions before images (i.e., emphasizing image generation conditioned on texts) 80\% of the time, and order the images before captions for the rest.

% We train a 86M parameter VAE following Esser et al. [2021].
% We use a CNN encoder and decoder, and latent dimension 8. The training objective is combines
% reconstruction and regularization losses.14 Our implementation reduces an image of 256×256 pixels
% to a 32×32×8 tensor, where each latent 8-dimensional latent pixel represents (conceptually) an 8×8

\paragraph{Model Details}
For image tokenization, we use the same VAE encoder\footnote{\url{https://huggingface.co/stabilityai/sd-vae-ft-mse}} as Transfusion to compress an image of $256 \times 256$ pixels into a $32 \times 32 \times 8$ tensor. These tensors are then passed into a 2-block U-Net downsampler \citep{ronneberger2015u} to further reduce  dimensions, resulting in a sequence of 256 patches (tokens). 
Both text-specific and image-specific Transformer modules are initialized from the pretrained Llama-3 8B model \citep{llama3}. 
% , a crucial design to reuse compute of pretrained LLMs. 
The U-Net downsampler and a corresponding U-Net upsampler are trained from scratch, together containing 0.27 billion parameters.
% The U-Net downsampler and a corresponding U-Net upsampler following the image Transformer outputs are trained from scratch, in total with 0.27B parameters. 
Like Transfusion, \name uses a maximum context length of 4096 tokens. 
% resulting in a total of 16.27B parameters (8B for text and 8B for image component, 0.27B for U-Net). 
% \swj{TODO: point to unet}.

\paragraph{Optimization}
In our main experiments, to preserve the language-only performance, we freeze the text modules ($\eta_\text{text}=0$) while training only the image modules using an AdamW optimizer ($\beta_1=0.9$, $\beta_2=0.95$, $\epsilon=1 \times 10^{-8}$) with a learning rate $\eta_\text{image}=1 \times 10^{-4}$. The learning rate follows a cosine decay schedule with a 4000-step warmup period before gradually decreasing to $1.5 \times 10^{-5}$.  

% in batches of 2M tokens. The training runs for 250K steps, processing approximately 0.5T tokens in total, with 256 H100 GPUs over XX days. 
% {\color{blue} [Han: 5 days? can't remember too well]}

\subsection{Controlled Comparison with Transfusion}
Our key model comparisons are with the original Transfusion 7B model \citep{transfusion},\footnote{Transfusion 7B and Llama-3 8B have the same Transformer sizes. The size difference is due to the different vocabularies, which affects input and output embedding layers.} 
which was trained for 250K steps on 0.25T language-only tokens (text data) and 0.25T image-captions tokens (image data).

Since we freeze the text module during training, we can exclude text data from our training process while maintaining language capabilities. This design choice allows us to explore two training configurations for a controlled comparison with Transfusion: 
In the first configuration, 
% (\textit{$0.5\times$image FLOPs}), 
we match the amount of 0.25T image data used by Transfusion while leaving out the text data. As a result, this variant of \name uses approximately half the total FLOPs of Transfusion. 
In the second configuration, 
% (\textit{$1\times$image FLOPs}), 
we match Transfusion by using the same total FLOPs,
% \footnote{Since our goal is to investigate continued training from pretrained LLMs we exclude the FLOPs of pretraining the text-only LLMs in the comparison.} doubling the amount of image data.
% \swj{are we able to map these configurations with the results in table 1. is it clear? }

Additionally, for the language-only tasks, we report the performance of Llama-3 8B model to demonstrate that our model is able to maintain its strong text performance. 

% As we freeze the text module, we don't need to train our model on text data, which significantly reduce 
% We train our model in two different configurations for fair comparison.
% using half of Transfusion's total batch size (

% the same total batch size (twice image data, no text data). 
% We perform controlled comparisons with Transfusion \citep{transfusion} with the same model size and datasets. We have two configurations of LlamaFusion, one with the same training FLOPs as Transfusion's text and image training, and one with the same FLOPs as Transfusion's image training only (half total FLOPs).  

% \swj{conduct extensive FLOPs-controlled experiments}

\subsection{Evaluation Setup} \label{sec: evalsetup}

Following Transfusion, we evaluate \name on language-only, image understanding, and image generation tasks. 

\paragraph{Language-only} We evaluate the model's language abilities using four tasks from the standard Llama evaluation suite \citep{llama3}, including Hellaswag \citep{zellers2019hellaswag}, PIQA \citep{bisk2020piqa}, SIQA \citep{sap2019social}, and WinoGrande \citep{sakaguchi2021winogrande}. We report accuracy on these benchmarks.
% \footnote{Since \name freezes the text module, its performance on additional language tasks would be same as Llama-3's results.} 

\paragraph{Image Generation} For evaluating image generation, we use the MS-COCO benchmark \citep{lin2014microsoft}. We generate images for 30K randomly selected prompts from the validation set and measure the Frechet Inception Distance (FID) \citep{heusel2017gans} and CLIP scores \citep{clip}. Our image generation results include versions obtained 
without classifier-free guidance (CFG coefficient of 1.0) and with a CFG coefficient of 1.55 or 1.6. 
% with a classifier-free guidance (CFG) coefficient of 1.0 
% for all models for a fair comparison. 
% \swj{may change it for 1.6}
% \lili{cfg=1.0 means, no cfg. Just say we don't use cfg in our ablation, and for the final model, we use cfg=1.6}

\paragraph{Image Understanding} We evaluate the models' ability to generate image descriptions 
using the  test split of MS-COCO \citep{lin2014microsoft}, reporting CIDEr scores \citep{vedantam2015cider}. 

% Karpathy

\subsection{Results} \label{sec: results}

% In \autoref{tab:main}, we show the performance of two variants of \name compared to Transfusion. 
\autoref{tab:main} compares two variants of \name against Transfusion. 
On language-only benchmarks, \name keeps the strong performance of Llama-3 since we freeze all text modules. 
% by setting $\eta_{\text{text}}=0$. 
For image understanding, \name substantially surpasses Transfusion, with a 20\% improvement. 
In image generation tasks, \name also shows superior results in both FID and CLIP scores. 

% On language understanding benchmarks, \name keeps text performance of LLaMA 3. Additionally, it significantly surpasses Transfusion in image captioning, with a 20\% improvement. In image generation tasks, \name shows superior results in both FID and CLIP scores.

\begin{figure*}[ht]
    \centering
    \includegraphics[width=\linewidth]{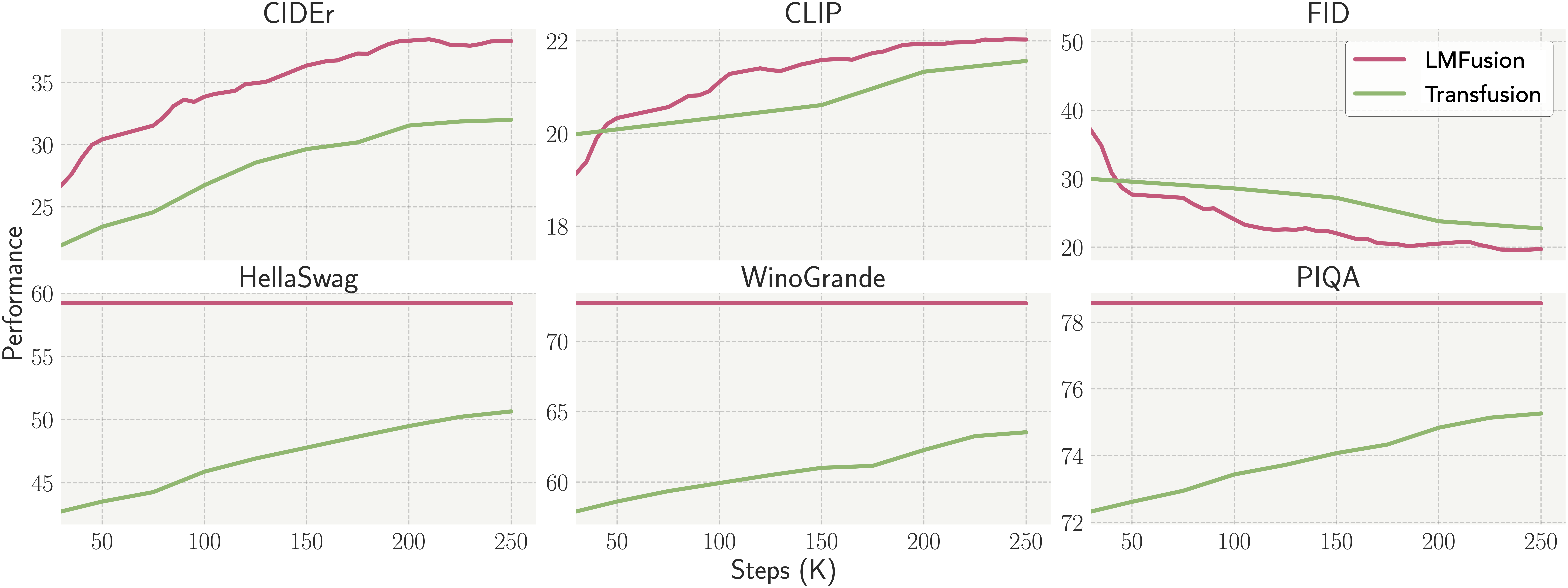}
    \caption{\textbf{Evaluation of \name and Transfusion during training.}
\name keeps the text performance of Llama throughout training, while achieving better image understanding ability (CIDEr) and image generation quality (CLIP, FID). 
    }
    \label{fig:performance_vs_token}
    \vspace{-1.0em}
\end{figure*}

In \autoref{fig:performance_vs_token}, we benchmark the performance of \name and Transfusion throughout the training.\footnote{For the image generation results plotted throughout the training, we use a smaller subset of 5K prompts and without classifier-free guidance.} 
% \name's text performance remains unchanged by freezing text modules.
We observe a consistent advantage of \name over Transfusion during the entire training, for image captioning and generation. These results suggest that \name effectively leverages the pretrained language modules from Llama while developing strong image abilities. 
% Although \name’s model size is twice that of Transfusion, it has the same FLOPs, as only one transformer’s parameters are active for each token. 
Although \name has twice as many parameters as Transfusion, it uses same FLOPs since only half of the parameters are activated for each input token from an arbitrary modality.  

\begin{table*}[t]
    \centering
    \begin{tabular}{lrcccccc}
        \toprule
        & & \multicolumn{3}{c}{\small \textbf{Language-only}} & \multicolumn{1}{c}{\small \textbf{Image}} & \multicolumn{2}{c}{\small \textbf{Image Generation}} \\
        & & \multicolumn{3}{c}{\small \textbf{Evaluation}} & \multicolumn{1}{c}{\small \textbf{Understanding}} & \multicolumn{2}{c}{\small 
        % \textbf{Generation} 
        {\scriptsize (\texttt{without} $\mid$ \texttt{with} CFG)}} \\
        \cmidrule(lr){3-5} \cmidrule(lr){6-6} \cmidrule(lr){7-8}
        Model & {\footnotesize Rel. FLOPs} & {\footnotesize HellaSwag $\uparrow$} &  {\footnotesize SIQA $\uparrow$} & {\footnotesize WinoGrande $\uparrow$} & {\small CIDEr $\uparrow$} & {\small FID $\downarrow$} & {\small CLIP $\uparrow$} \\
        \midrule
        Llama 3 & - & 60.0 &  48.1 & 72.8 & -- & -- & -- \\
        \midrule
        Transfusion &  1$\times$  & 51.0 & 42.3 & 64.3 & 32.0 & 14.4 $\mid$ 8.70 & 22.1 $\mid$ 24.4 \\

         \name{} & 0.5$\times$ & 60.0 &  48.1 & 72.8 & 38.3 & 13.9 $\mid$ 8.75 & 22.0 $\mid$ 24.4 \\
        \name{} & 1$\times$ & 60.0 &  48.1 & 72.8 & 38.4 & 14.0 $\mid$ 8.61 & 22.1 $\mid$ 24.4 \\
        \bottomrule
    \end{tabular}
    \caption{\textbf{Results across text-only benchmarks, image understanding and image generation.} \name{} preserves Llama-3's text performance while adding strong image understanding and generation capabilities. Using only half of the total training FLOPs, it outperforms Transfusion across all tasks, with particularly notable improvements in image understanding and text benchmarks, thanks to its initialization from Llama-3. Image generation results are obtained without classifier-free guidance (CFG) or with a CFG factor of 1.55. 
    % \swj{final fid/clip performance} 
    % \swj{not sure if flops discusion here is clear. we don't really explain what is $0.5\times$image + $0.5\times$text FLOPs in text. would it confuse people?} \lili{can you just use relative flops, transfusion is 1X, and the two settings are 0.5X and 1X; add a column, instead of extend the model name}
    % \han{new secondary FID/CLIP: cfg1.55, on 30k data}
    }
    \label{tab:main}
    \vspace{-0.5em}
\end{table*}

\begin{table*}[t]
    \centering
    \begin{tabular}{l c c c c c c}
        \toprule
        & & \multicolumn{4}{c}{\small \textbf{Image Understanding}} & \multicolumn{1}{c}{\small \textbf{Image Generation}} \\
        \cmidrule(lr){3-6} \cmidrule(lr){7-7}
        Model & {\footnotesize Base LLM} & {\footnotesize MMMU $\uparrow$} & {\footnotesize ChartQA $\uparrow$} & {\footnotesize RealWorldQA $\uparrow$} & {\footnotesize MME-P $\uparrow$} & {\small FID $\downarrow$} \\
        \midrule
        EMU-3  & -- & 31.6 & 51.8 & 57.4 & -- & 12.8 \\
        Show-O  & Phil-1.5 1.3B & 27.4 & -- & -- & 1435.7 & 9.2 \\
        Janus  & DeepSeek 1.3B & 30.5 & -- & -- & 1338.0 & 8.5 \\
        Chameleon  & -- & 28.4 & 0.0 & 19.6 & -- & 26.7 \\
        MetaMorph & LLaMA-3.1 8B & \textbf{41.8} & 37.1 & 58.3 & -- & 11.8 \\
        Transfusion & -- & -- & -- & -- & -- & 8.7 \\
        \midrule
        LLaVAFusion & LLaVA-Next 8B & 41.7 & \textbf{69.5} & \textbf{60.0} & \textbf{1603.7} & \textbf{8.2} \\
        \bottomrule
    \end{tabular}
    \caption{\textbf{Comparison of multimodal models across image understanding and generation capabilities.} Models are evaluated on various image understanding benchmarks and image generation quality (FID). The models without base LLM are pretrained from scratch.} \label{tab:imagenew}
    \label{tab:comparison}
\end{table*}
% include the figs/performance_vs_token.pdf, make it page width

% Evaluation of OLMOE-1B-7B and the current best OLMo models during pretraining. OLMOE-1B-7B differs from the OLMo models in its MoE architecture, several training hyperparameters, and its training dataset, see §2. A version
% of this plot with tokens as the x-axis and markers where annealing starts is in Appendix E.

\begin{figure*}[ht]
    \centering
    \includegraphics[width=\linewidth]{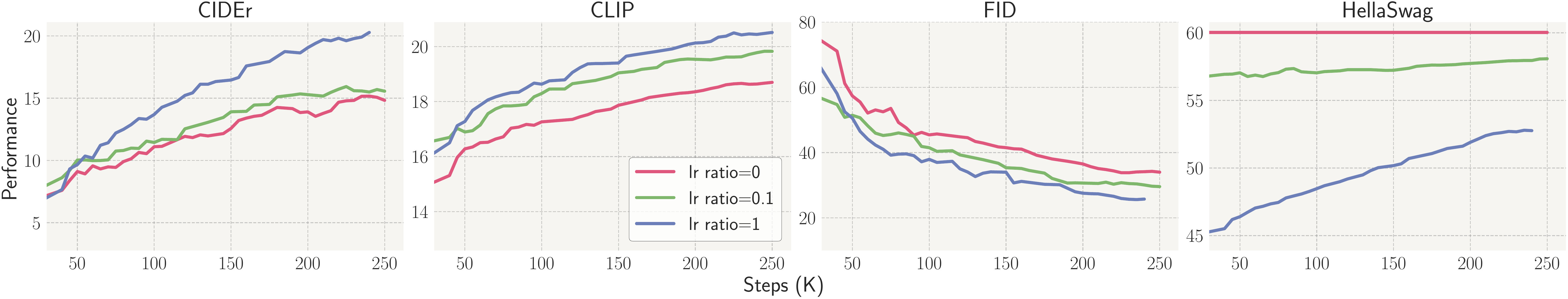}
    \caption{\textbf{Performance of naive Llama-3 finetuning (no separation) with varying lr ratio $\frac{\eta_{\text{text}}}{\eta_{\text{image}}}$. }
    When directly finetuning the Llama-3 model for multimodal generation, using the same learning rate for both text and image components (lr ratio $=$ 1) substantially reduces its text-only performance. Lowering the learning rate for the text component relative to the image component (lr ratio $<$ 1) helps preserve language performance but slows down the acquisition of multimodal abilities. 
    % \lili{why lr=1, 250k step is missing? }
    }
    \label{fig:ablation_dense_performance}
    \vspace{-0.5em}
\end{figure*}

\begin{figure*}[ht]
    \centering
    \includegraphics[width=\linewidth]{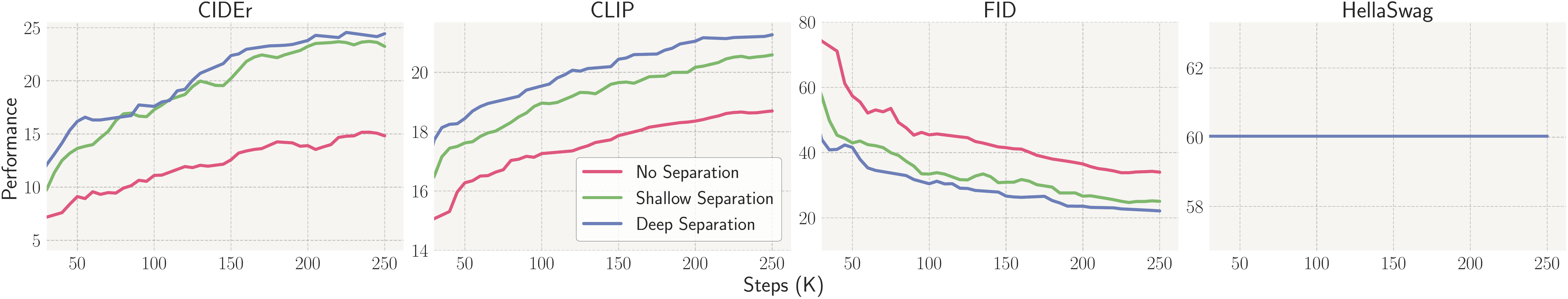}
    \caption{\textbf{Performance of \textit{no separation} (dense model), \textit{shallow separation }(modality-specific FFNs only), and \textit{deep separation }(modality-specific FFNs and attention) when text modules are frozen. }
    Deep modality separation outperforms shallow separation and no separation. 
    % While both separation approaches significantly outperform the dense model, deep separation achieves the best results, particularly on image generation tasks. All models preserve the original text capabilities since the text components are frozen.
    }
    \label{fig:ablation_dense_moe_mot_performance}
    \vspace{-0.5em}
\end{figure*}
% We compare three architectural approaches with frozen text components (lr ratio $\frac{\eta_{\text{text}}}{\eta_{\text{image}}} = 0$): no separation (dense model), shallow separation (modality-specific FFNs only), and deep separation (modality-specific FFNs and attention).

\begin{figure*}[ht]
    \centering
    \includegraphics[width=\linewidth]{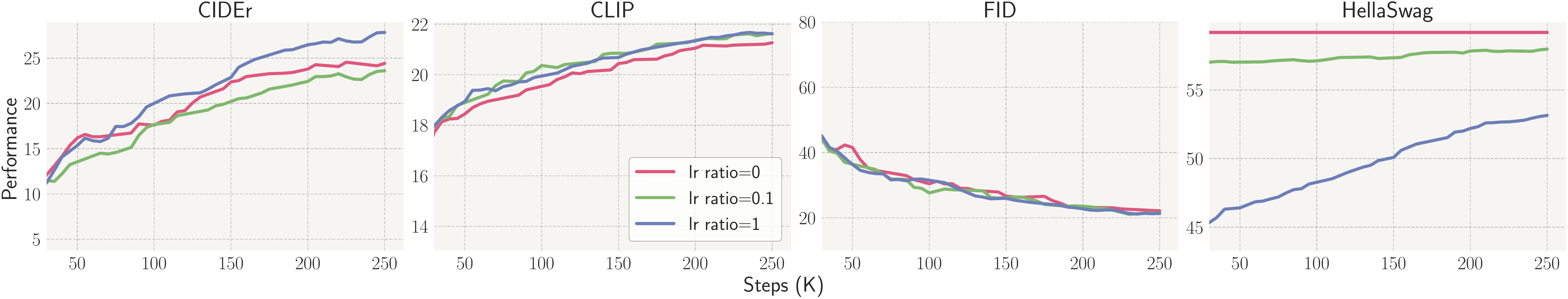}
    \caption{\textbf{Performance of deep modality separation with varying lr ratios $\frac{\eta_{\text{text}}}{\eta_{\text{image}}}$.} When the text modules are frozen (lr ratio = 0), deep separation preserves language capabilities and performs strongly on both image understanding and generation, unlike the dense models.}
        \label{fig:ablation_mot_performance}
        \vspace{-0.5em}
\end{figure*}

\section{Analysis}
\label{sec:analysis}
Central to \name is our modality separation techniques, which employs the design of modality-specific modules and decoupled learning rates for language and image modules. Our architectural ablation (\S \ref{sec: ablation}) demonstrates the importance of the design for maintaining model performance across both modalities. Additionally, we showcase \name's ability to generalize to image-to-image generation through image editing tasks, which require simultaneous understanding of both input images and textual prompts (\S \ref{subsec: edit}). We further showcase that this recipe could be used for adapting  

% \name introduces image generation and understanding capabilities to a pretrained text-only LLM through a Transfusion training that separates language and image modalities with decoupled learning rates. 

% Finally, we also demonstrate \name's ability to generalize to image-to-image generation through image editing tasks, which require simultaneous understanding of both input images and textual prompts (\S \ref{subsec: edit}). 

\subsection{Architecture Ablations} \label{sec: ablation}
% Our analysis (\S \ref{subsec: design}) shows the importance of this architectural decision: First, we show that directly tuning the original LLaMA model on multimodal data without architectural modifications results in significant performance degradation on language-only benchmarks. 

% We find learning rate decoupling can mitigate the language-only performance degradation but also compromises image-related tasks. 
% We then examine the effects of implementing varying degrees of modality separation and find deeper modality separation is more beneficial to the multimodal continued training. 
% Finally, we also demonstrate \name's ability to generalize to image-to-image generation through image editing tasks, which require simultaneous understanding of both input images and textual prompts (\S \ref{subsec: edit}). 

% The key innovation lies in our modality separation strategy, which employs (1) learning separation: decoupled learning rates for text and image modules (2) architectural separation: separate image and text modalities..

% \lili{are all the lr=0 experiments in 4,5,6 without text data.?} 
% \han{in all ablation cases, training data still includes text data for a fair comparison (but lr=0 effectively ignores them)} 
\subsubsection{Experimental Design}
To evaluate different design choices, we conduct ablation studies using small-scale variants of \name. Our analysis focuses on the impact of modality separation by comparing three designs: (1) \textit{no separation} (a single dense model), (2) \textit{shallow separation} (using modality-specific FFNs only), and (3) \textit{deep separation} (using both modality-specific FFNs and attention mechanisms, our final \name).

% Throughout our experiments, we examine the impact of learning rate dynamics by varying the ratio between text and image learning rates ($\frac{\eta_{\text{text}}}{\eta_{\text{image}}}$) across three settings: {0, 0.1, 1}, with a constant image learning rate $\eta_{\text{image}}=1 \times 10^{-4}$. A ratio of 1 represents standard continual pretraining where all components share the same learning rate, while a ratio of 0 indicates complete freezing of text-related components.

% To investigate architecture design ablations, we train small-scale \name with different variants. Specifically, we examine the necessity of modality separation by comparing the design of no separation (one dense model without modality-specific moduels), shallow separation (only modality-specific FFNs) and deep seoaration (modality-specific FFNs and attention, final \name). 

% In our ablation study, we mainly aim to investigate whether the modality-specific modules (modality separation) and learning rate decoupling are necessary. 

\paragraph{No separation (dense model)}

We begin our experiments with the dense Llama-3 8B model trained using the Transfusion recipe. This dense model maintains a unified structure where most components are shared across modalities (a single set of $\operatorname{QKV}$, $\operatorname{O}$ and $\operatorname{FFN}$ process both texts and images), with the exception of U-Net upsampler and downsampler. 
% For training, we implement a \textbf{learning rate decoupling strategy}. 
For training, we use a text learning rate ($\eta_{\text{text}}$) for the components initialized from the text-only LLM \{\blue{
$\operatorname{Proj_{\text{text}}}$,
$\operatorname{QKV}$,
$\operatorname{O}$,
$\operatorname{FFN}$,
$\operatorname{LM-Head}_{\text{text}} $
}\}, and an image learning rate $\eta_{\text{img}}$ for \{\orange{
$ \operatorname{UNet-Down}_{\text{img}}$,
$\operatorname{UNet-Up}_{\text{img}}$}\}. 
To investigate the impact of learning rate decoupling, we experiment with various learning rate ratios $\frac{\eta_{\text{text}}}{\eta_{\text{image}}} \in \{ 0, 0.1, 1 \}$, with a constant image learning rate $\eta_{\text{image}}=1 \times 10^{-4}$, the same as the main experiments. A ratio of 1 represents standard continual pretraining where all components share the same learning rate, while a ratio of 0 indicates a complete freezing of text-related components.

\paragraph{Shallow separation (modality-specific FFNs only)}
We explore a simplified variant of \name that separates only FFNs into text-specific and image-specific modules---a common approach in mixture-of-experts architectures \citep{lin2024momaefficientearlyfusionpretraining, muennighoff2024olmoe}. In this setup, we use a single shared attention mechanism ($\operatorname{QKV}$ , $\operatorname{O}$) for processing both image and text data. 
For training, we employ separate learning rates: $\eta_{\text{text}}$ for text-related components \{\blue{
$\operatorname{Proj}_{\text{text}}$,
$\operatorname{QKV}$,
$\operatorname{O}$,
$\operatorname{FFN}_{\text{text}}$,
$\operatorname{LM-Head}_{\text{text}} $
}\} and $\eta_{\text{img}}$ for image-related components \{\orange{
$ \operatorname{Unet-Down}_{\text{img}}$,
$\operatorname{FFN}_{\text{img}}$,
$\operatorname{Unet-Up}_{\text{img}}$}\}. We experiment with various learning rate ratios $\frac{\eta_{\text{text}}}{\eta_{\text{image}}} \in \{ 0, 0.1, 1 \}$.

% only \blue{$\operatorname{FFN}_\text{text}$} and \orange{$\operatorname{FFN}_\text{image}$} are initialized from the original Llama-3 \blue{$\operatorname{FFN}_\text{orig}$}, while all other Transformer modules retain their Llama-3 initialization.

% An alternative approach to \name is to only separate text-oriented and image-oriented FFNs. This is a canonical setup in the MoE literature. 
% I.e., only \blue{$\operatorname{FFN}_\text{text}$} and \orange{$\operatorname{FFN}_\text{image}$} are reduced to \blue{$\operatorname{FFN}_\text{orig}$}. 
% All Transformer modules are initialized from Llama-3. 
% We apply a text learning rate, $\eta_{\text{text}}$, for \{\blue{
% $\operatorname{Proj}_{\text{text}}$,
% $\operatorname{QKV}_\text{orig}$,
% $\operatorname{O}_{\text{orig}}$,
% $\operatorname{FFN}_{\text{text}}$,
% $\operatorname{LM-Head}_{\text{text}} $
% }\}
% , and an image learning rate, $\eta_{\text{img}}$, for \{\orange{
% $ \operatorname{U-Down}_{\text{img}}$,
% $\operatorname{FFN}_{\text{img}}$,
% $\operatorname{U-Up}_{\text{img}}$}\}. 
% We set the learning rate ratio, $\frac{\eta_{\text{text}}}{\eta_{\text{image}}} = \{ 0, 0.1, 1 \}$. 

\paragraph{Deep separation (modality-specific FFNs and attention)} 
Our \name, as described in \autoref{sec:method}, represents a deep separation design where both FFNs and attention mechanisms are modality-specific. While our primary configuration freezes text modules during training, we also analyze the impact of different learning dynamics by varying the learning rate ratio $\frac{\eta_{\text{text}}}{\eta_{\text{image}}}$ across $\{{0, 0.1, 1}\}$.

% Our \name described in \autoref{sec:method} is an example of 
% the deep separation design.  
% Apart from training with frozen text modules, we also set the learning rate ratio, $\frac{\eta_{\text{text}}}{\eta_{\text{image}}} = \{ 0, 0.1, 1 \}$ for analysis. 

\medskip

% Throughout the analysis, we train a smaller-scale \name along with the ablated dense model and the model with shallow separation. 
% In ablation study, all models are trained  for 250K steps with input sequences of 4096 tokens in batches of 250K tokens, processing 0.03T text-only tokens (text data) and 0.03T image-text tokens (image data).  All other hyperparameters remain the same as in the main experiments. 
In the ablation study, all models are trained for 
250K training steps with a sequence length of 4,096 tokens and a batch size of 250K tokens.
The training data comprised 0.03T text-only tokens and 0.03T image-caption tokens. All other hyperparameters remained consistent with those employed in our main experiments.

% 250K steps with input sequences of 4096 tokens in batches totaling 250K tokens, processing 0.03T text-only tokens (text data) and 0.03T image-text tokens (image data). All other hyperparameters match those used in the main experiments.

\subsubsection{Results}
\vspace{-2mm}

% \swj{does some part of analysis sound bit repetitive? }
\paragraph{\textbf{Naive finetuning of dense pretrained LLMs for multimodal generation compromises their original language capabilities.}} 
% \swj{directly adapting clear?}
When directly finetuning Llama-8B (no separation) using the Transfusion recipe, we observe significant performance trade-offs between image and text capabilities (\autoref{fig:ablation_dense_performance}). 
% As shown in \autoref{fig:ablation_dense_performance}, 
% in the standard continual training setup where all components use the same learning rate (lr ratio $\frac{\eta_{\text{text}}}{\eta_{\text{image}}} = 1$), 
With equal learning rates for text and image components ($\frac{\eta_{\text{text}}}{\eta_{\text{image}}} = 1$), the model shows continuous improvement in image understanding and generation. However, this comes at a substantial cost to language capabilities, with performance on HellaSwag dropping by 15\% initially. While language performance improves during training, it never recovers to the original Llama-3 model's level, maintaining a persistent 7\% gap.

% \vspace{-2mm}
To mitigate this issue, we explore setting $\frac{\eta_{\text{text}}}{\eta_{\text{image}}} < 1$, which allows us to train image-specific modules (U-Nets) with a regular learning rate while preserving text capabilities using a smaller learning rate for the general Transformer components. \autoref{fig:ablation_dense_performance} shows this improves language-only benchmark performance, reducing the gap from 7\% to 2\% when the ratio is 0.1.
% $\frac{\eta_{\text{text}}}{\eta_{\text{image}}} = 0.1$
 However, for dense models, this improvement comes at the cost of consistently reduced image capabilities. Overall, while learning rate decoupling offers some mitigation to the text performance drop, training dense pretrained LLMs without modality separation remains suboptimal. 

% By setting $\frac{\eta_{\text{text}}}{\eta_{\text{image}}} < 1$, we aim to train the image-specific modules (U-Nets in the dense model) more with a regular learning rate and retain the capability of the pretrained text-specific modules with a significantly smaller learning rate. 
% We find such learning rate decoupling is helpful to the language-only benchmark performance, reducing the performance gap to the original LLM from 7 to 2 points when $\frac{\eta_{\text{text}}}{\eta_{\text{image}}} = 0.1$. 
% However, in the case of dense models, this also compromises the image capabilities consistently. 
% Overall, continuing training a dense pretrained LLM without modality separation leads to suboptimal performance though with learning rate decoupling as mitigation. 

% While language performance improve during training, it remains significantly below the original pretrained Llama-3 model's capability, showing a persistent 7\% gap.

% In \autoref{fig:ablation_dense_performance}, we show the results of continuing training a dense Llama-3 model with the Transfusion recipe. 
% We first observe in a vanilla setup where $\frac{\eta_{\text{text}}}{\eta_{\text{image}}} = 1$, the image understanding and generation performance keeps growing throughout the training. However, on the language-only benchmark HellaSwag, the performance drops a substantial 15 points from the beginning. Though the language-only performance also increases during training, it never recovers near the level of our initializing pretrained Llama-3 model, with a large gap of 7 points. 
\captionsetup[subfigure]{labelformat=empty}

\begin{figure*}[tp]
    \centering
    \subfloat[Change the fork to a spoon.]{\includegraphics[width=0.3\textwidth]{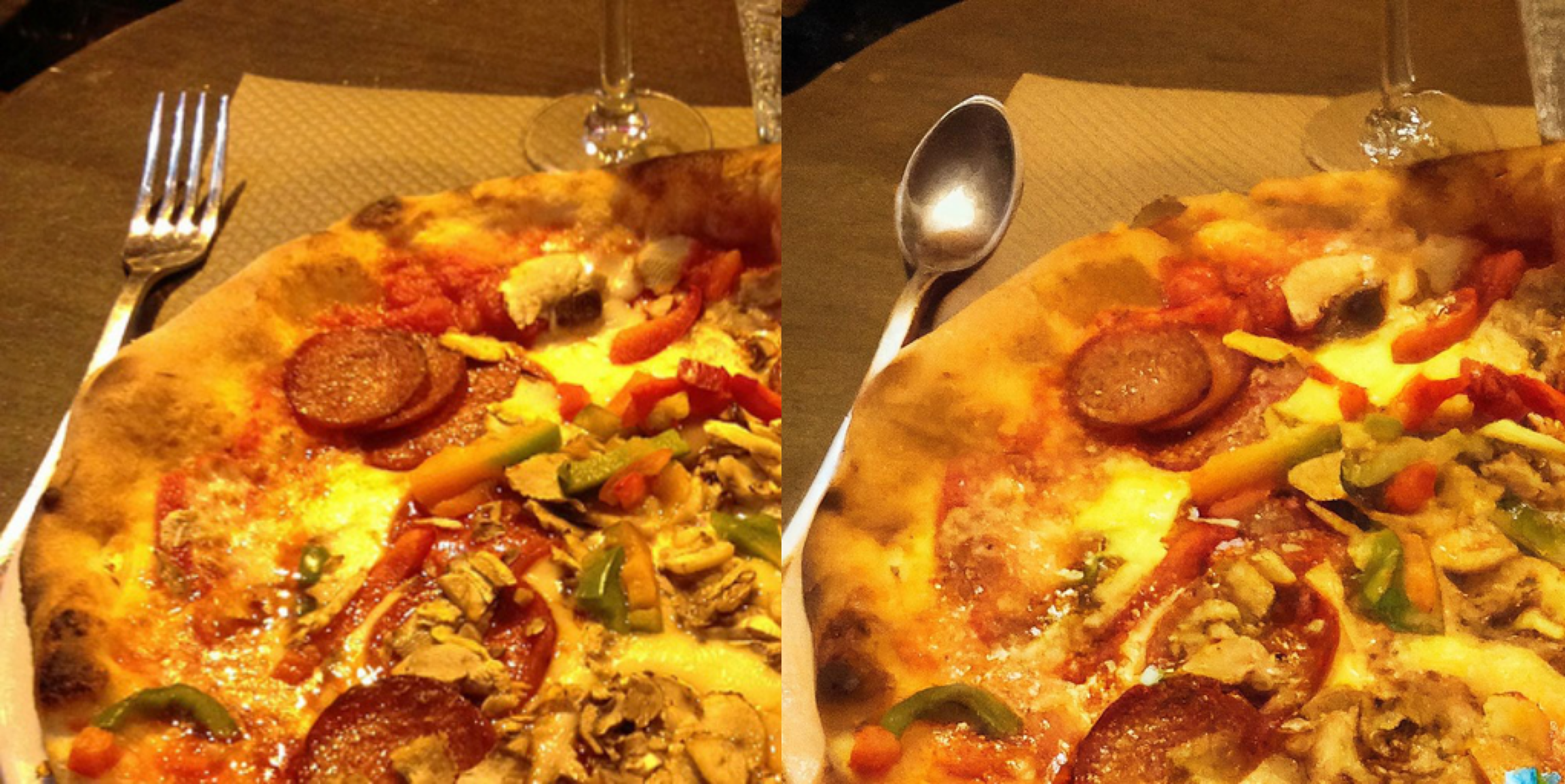}} \hspace{0.03\textwidth} 
    % \subfloat[Let the laptop screen be blank. \swj{change the image}]{\includegraphics[width=0.38\textwidth]{figs/edit/pc_combined.png}}\\
    % \centering
    \subfloat[Add a soda can in the back.]{\includegraphics[width=0.3\textwidth]{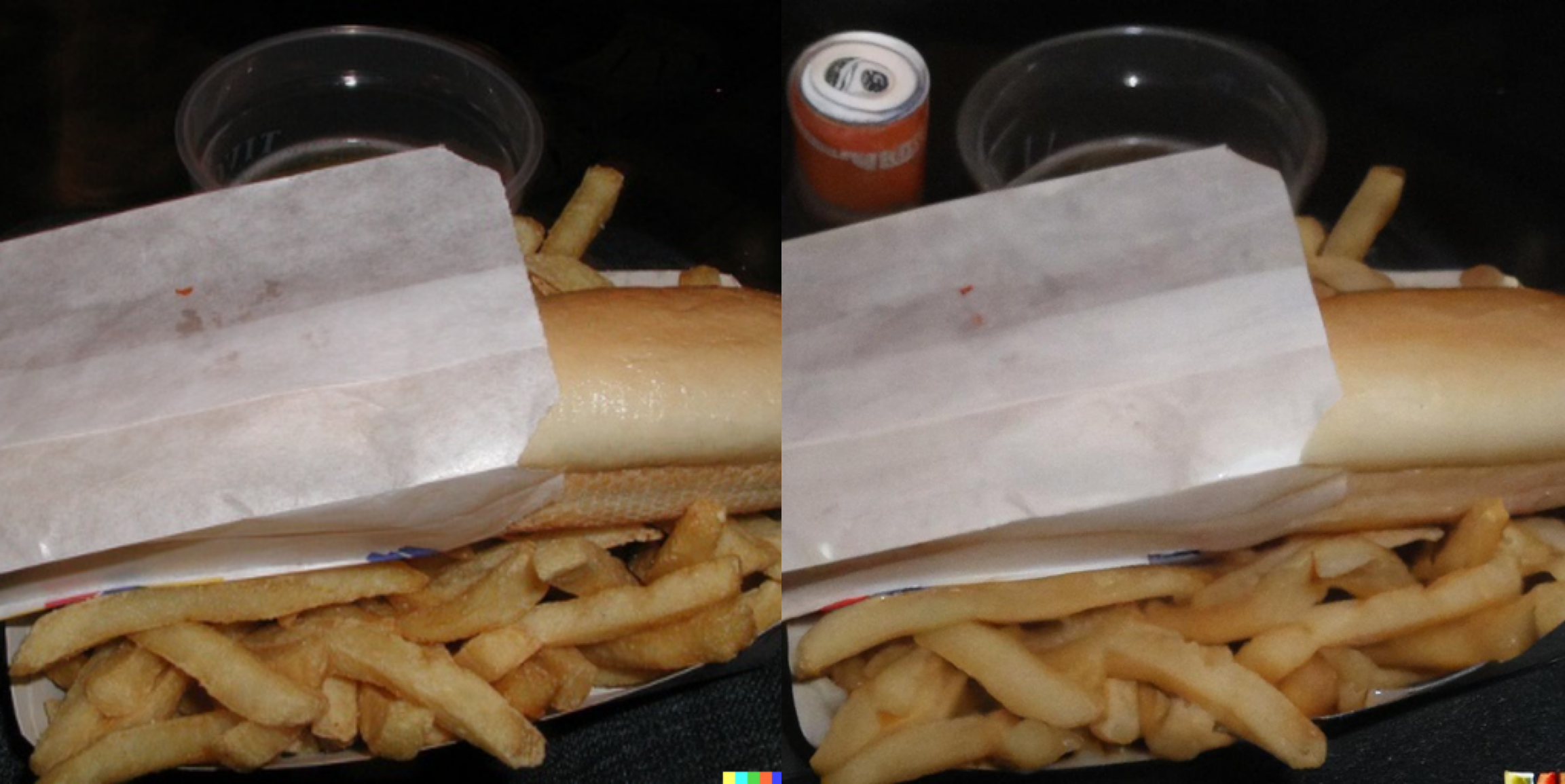}} \hspace{0.03\textwidth}
    \subfloat[Let there be a painting instead of a sign.]{\includegraphics[width=0.3\textwidth]{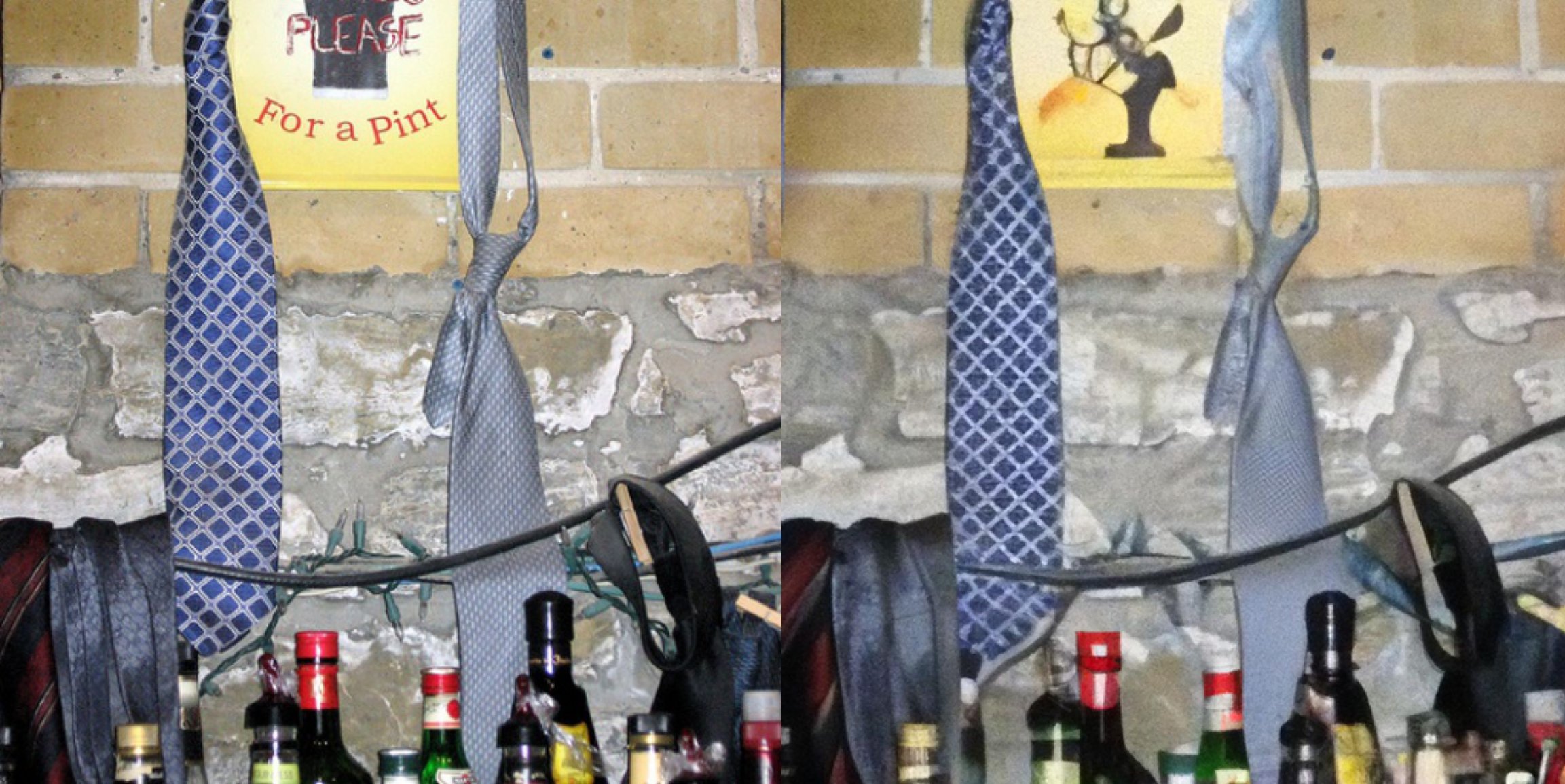}}\\

    % \subfloat[Add a blue rug to the floor.]{\includegraphics[width=0.23\textwidth]{img_edit_samples/0_add_blue_rug.jpg}\includegraphics[width=0.23\textwidth]{img_edit_samples/1_add_blue_rug.jpg}}\hfill
    % \subfloat[Delete the overhead lights on top of the sink.]{\includegraphics[width=0.23\textwidth]{img_edit_samples/0_remove_light.jpg}\includegraphics[width=0.23\textwidth]{img_edit_samples/2_remove_light.jpg}}\\[10pt]
    
    % \subfloat[Change the roll of thread into a roll of wire.]{\includegraphics[width=0.23\textwidth]{img_edit_samples/0_thread_to_wire.jpg}\includegraphics[width=0.23\textwidth]{img_edit_samples/1_thread_to_wire.jpg}}\hfill
    % \subfloat[Change the baseball bat to all brown.]{\includegraphics[width=0.23\textwidth]{img_edit_samples/0_change_bat_to_brown.jpg}\includegraphics[width=0.23\textwidth]{img_edit_samples/1_change_bat_to_brown.jpg}}\\[10pt]
\caption{\textbf{Edited images from a finetuned \name model.}}
\label{fig:edit2}
\vspace{-1.0em}
\end{figure*}

\paragraph{\textbf{Deep Modality Separation Outperforms Shallow Separation.}}
In \autoref{fig:ablation_dense_moe_mot_performance}, we compare three architectures: no separation (dense), shallow separation (modality-specific FFNs only), and deep separation (modality-specific FFNs and attention). We set $\frac{\eta_{\text{text}}}{\eta_{\text{image}}} = 0$ (freezing the text module) across all models to maintain Llama-3's text performance. Both separation approaches significantly outperform the dense model on all image benchmarks. While shallow separation performs slightly worse on image understanding, the performance gap widens notably in image generation tasks. 
% \lili{deep seperation with freezing text and shallow separation has the same number of tunable parameters. And it is still better than no separation lr=1, you can discuss this} 
% \han{good point! (a potential reason is that though they have the same amount of tunable parameters, the no separation case needs to spend part of them for pure texts, indicating the existence of knowledge conflicts)}  

Additionally, deep separation with $\frac{\eta_{\text{text}}}{\eta_{\text{image}}} = 0$ has the same amount of \emph{tunable} parameters as no separation with $\frac{\eta_{\text{text}}}{\eta_{\text{image}}} = 1$. 
Despite the intrinsic advantage of modality separation for text-only tasks, for image understanding and generation, 
we still observe that deep separation (blue curve in \autoref{fig:ablation_dense_moe_mot_performance}) are better than no separation (blue curve in \autoref{fig:ablation_dense_performance}). 
These results demonstrate that modality separation is crucial for effectively adapting pretrained language-only LLMs for multimodal generation. 

\vspace{-2mm}
\paragraph{\textbf{Analyzing learning rate decoupling strategy w.r.t. modality separation.}} 
% \paragraph{Freezing the text module in \name leads to strong multimodal generation while preserving its original language capabilities.}
The impact of freezing text modules varies dramatically between architectures. In dense models (\autoref{fig:ablation_dense_performance}), freezing text components  
($\frac{\eta_{\text{text}}}{\eta_{\text{image}}} = 0$)
significantly impairs both image understanding and generation compared to full fine-tuning. However, in the deep modality separation setting shown in \autoref{fig:ablation_mot_performance}, freezing the text module not only maintains the original text performance but achieves strong performance on image understanding and generation, unlike the dense models.  
% \lili{the title is overclaiming. I think you can just objectively describe what happens when we combine deep separation and "learning rate decoupling strategy". And comparing lr=0, 0.1, 1. We still think lr=0 is the best on overall performance. And it also only use the half of the flops }

% As shown in \autoref{fig:ablation_dense_performance}, in dense models without modality separation, freezing text-related components ($\frac{\eta_{\text{text}}}{\eta_{\text{image}}} = 0$) substantially degrades performance on image understanding and generation compared to no freezing ($\frac{\eta_{\text{text}}}{\eta_{\text{image}}} = 1$). 

% In analysis, we show why this design choice is important \autoref{subsec: design}. We first show that directly finetuning original LLaMA model on the multimodal data without architecture modificaiton leads to significant drop on the language understanding benchmarks. Then we show different levels of separation. 
% Lastly, we show \name is able to perform image editig , an image-to-image generaiton that requires models understanding of input image nad textual prompt. 
% generate images based on multimodal input such as  

% \subsection{Model design ablation} \label{subsec: design}
% \subsubsection{}
% \begin{enumerate}
%     \item four figures in a row: cider, fid, 2 text
%     \item directly finetunes the dense model fails on text performance (winogrande and piqa)
%     \item Different learning rates for (language) transformer and (image) UNet tradeoff...
%     \item No separation vs. shallow (MoE) separation vs. deep (MoT) separation.
% \end{enumerate}

\subsection{Image editing} \label{subsec: edit}
\name, our unified multimodal generative model, is naturally well-suited for tasks involving interleaved data types, such as image editing. Following Transfusion, we finetune \name on the same dataset of 8K image editing examples, each consisting of an original input image, a prompt detailing the desired edit, and a resulting image that reflects the specified changes. In \autoref{fig:edit2}, we apply the finetuned \name to input images and editing prompts from the MagicBrush \citep{zhang2024magicbrush} test set. Qualitative results demonstrate that \name performs effectively in these image-editing scenarios, complementing its strong capabilities in text-only, image understanding, and image generation tasks.

\subsection{LLaVAFusion: extending \name to vision-language models}

\name continues training the language-only pretrained LLM Llama with the Transfusion recipe. 
Can this 
% Transfusion-style continued training work 
recipe be extended to
on vision-language models (VLMs) such as LLaVA \citep{liu2024visual, liu2024llavanext} and Qwen-VL \citep{bai2023qwen} as well? 
In this section, we extend the recipe of \name to VLMs, preserving their multimodal understanding capabilities while introducing image generation abilities. 
Specifically, we build on LLaVA-NeXT \citep{liu2024llavanext}, freezing its transformer parameters and integrating a dedicated, image-specific transformer module trained in parallel. We use the same data and model settings as \name. 
We refer to this new model as LLaVAFusion and demonstrate its image understanding performance on MMMU \citep{yue2023mmmu}, 
% MMBench \citep{liu2025mmbench}, 
MME-Perception \citep{fu2024mmecomprehensiveevaluationbenchmark}, 
ChartQA \citep{masry-etal-2022-chartqa}, and RealWorldQA\footnote{\url{huggingface.co/datasets/xai-org/RealworldQA}}, as well as its image generation results.
For baselines, we compare LLaVAFusion against EMU-3 \citep{wang2024emu3nexttokenpredictionneed}, Show-O \citep{xie2024showosingletransformerunify}, Janus \citep{wu2024janusdecouplingvisualencoding}, Chameleon \citep{chameleonteam2024chameleonmixedmodalearlyfusionfoundation}, MetaMorph \citep{tong2024metamorphmultimodalunderstandinggeneration}, and Transfusion \citep{transfusion}.
As shown in \autoref{tab:imagenew}, LLaVAFusion LLaVAFusion demonstrates strong performance in both image understanding and generation when compared to other unified multimodal LMs.
% maintains the image understanding performance of LLaVA-NeXT while achieving competitive results in image generation. 
This demonstrates that \name is promising as an extension not only to language-only LLMs but also to VLMs, enhancing the multimodal generation capabilities in both cases.

\section{Related Work}
\label{sec:related}
\paragraph{Unified Models for Multimodal Generation}
Recent work has extensively explored unified frameworks for multimodal generation, including text generation, image understanding, and image generation.  While texts are commonly represented as discrete tokens across models, approaches to representing images—especially for image generation—vary significantly. For instance, methods in \citep{lu2022unified,yu2023scaling,lu2024unified,team2024chameleon,xie2024show,wu2024vila, aiello2023jointlytraininglargeautoregressive}, represents images using vector-quantized discrete tokens \citep{van2017neural,esser2021taming,lee2022autoregressive}. An alternative method, adopted by \citep{sun2024generative,ge2024seed}, employs continuous embeddings that require a separate diffusion model for decoding.  In this work, we build upon Transfusion \citep{transfusion}, which integrates autoregressive generation for texts with diffusion for images within a single, end-to-end model.

% Despite representing texts similarly as discrete tokens, prominent unifying paradigms differ in their ways of representing images, especially in image generation. 
% For example, \citep{lu2022unified,yu2023scaling,lu2024unified,team2024chameleon,xie2024show,wu2024vila} represent images with vector-quantized, discrete tokens \citep{van2017neural,esser2021taming,lee2022autoregressive}. 
% \citep{sun2024generative,ge2024seed} model images with continuous embeddings that would depend on a separate diffusion model to decode. 
% In this work, we particularly build upon a seminal paradigm of Transfusion \citep{transfusion} that runs autoregressive generation on texts and diffusion on images, in one single, end-to-end model. 

% [VQ, Diffusion, ...]
% For example, Unified IO (1/2), Show-O, Chameleon, vila-u. 
% (Emu2, seed-x). 

% Model sparsity such as Mixture of Experts (MoE) \citep{} have shown been effective in making models more 
% Recently, 
% Inspired by the sparse models in LLM research, multimodal models are adopting the sparse designs as well when processing different modalities in one model. 
% Especially, there is a trend to separately process the data from different modalities, as they may have conflict in one set of model weights \citep{lyle2024switching}. For example, 
% \citep{shen2023scaling}
% For example, 
% 1. Moma
% 2. VLMo (Bao et al., 2022),
% BEiT-3 (Wang et al., 2022a) and VL-MoE (Shen et al., 2023)
% 3. MARS (https://arxiv.org/abs/2407.07614), 
% 4. (concurrent) MoT
% 5. playground v3 
\vspace{-2mm}
\paragraph{Model Sparsity} Model sparsity through Mixture of Experts (MoE) \citep{shazeer2017outrageously,muennighoff2024olmoe,fedus2022switch,lepikhin2020gshard} has proven highly effective in improving LLM training efficiency. This approach has recently been extended to multimodal models \citep{shen2023scaling,lyle2024switching,Lin2024MoELLaVAMO,he2024marsmixtureautoregressivemodels}, particularly to address potential conflicts between different modalities. For example, recent efforts \citep{Chen2023EVEEV,lin2024momaefficientearlyfusionpretraining,Wang2021VLMoUV,wang2022image} replace standard Transformer FFNs with modality-specific experts, enabling separate processing paths for different modalities. Our work takes this concept further by using modality-specific attention mechanisms. Concurrent work \citep{liu2024playgroundv3improvingtexttoimage,liang2024mixtureoftransformers} demonstrates the effectiveness of this deeper separation in multimodal pretraining and image generation. 
% \han{further discuss MoT here} 
\vspace{-2mm}
\paragraph{Reuse of LLMs in Multimodal Training}
Based on the strong language capabilities of LLMs, some recent models on multimodal generation initializes their models from pretrained, language-only LLMs. 
For example, \citep{ge2023making,sun2023generative,dong2023dreamllm,xie2024show,wu2024vila,he2024mars} continued training upon the weights of language-only LLMs \citep{touvron2023llama} or vision LLMs without visual generation capabilities \citep{bai2023qwen}.  
The main focus of our work is to effectively reuse 
% and leverage
pretrained LLMs for multimodal generation, particularly with the Transfusion recipe, without any compromise on the LLMs' existing text-only capabilities.\footnote{Concurrent to our work, \cite{liu2024playgroundv3improvingtexttoimage} tackles multimodal generation via a joint attention mechanism between a DiT structure \citep{peebles2023scalable} for images and a frozen Llama-3 \citep{llama3} for texts.}  
% \swj{ask Han}

\section{Conclusion}
\label{sec:conclusion}
We present \name, a framework designed to equip LLMs with multimodal generative capabilities. By using Llama-3 for text generation and integrating parallel transformer modules for image diffusion, \name efficiently reuses compute invested in pretrained LLMs. 
% This modular design opens new research possibilities. 
%  \name enables independent developments of language and vision modules, de-risking the complexities associated with a large-scale, joint multimodal pretraining. This architecture also provides a foundation for future extensions to other modalities like video and audio.

\name's modular design enables independent developments of language and vision modules, de-risking the complexities associated with a large-scale, joint-modality pretraining. 
% This architecture also provides a foundation for future extensions to other modalities like video and audio. 
While \name is currently built upon text-only LLMs, it can benefit further from existing visual understanding LLMs \citep{liu2023llava, dai2023instructblip, Liu_2024_CVPR, zhu2024minigpt}, inheriting the strong multimodal understanding ability while enabling generating interleaved text and visual content. 
% Furthermore, \name complements existing multimodal understanding models . While these models excel at understanding images, they currently lack generative capabilities. \name bridges this gap, enabling generation of interleaved text and visual content. 

% Its modality-separation architecture enables researchers to independently develop specialized modules for additional modalities, such as audio and video, which can then be seamlessly integrated into a unified multimodal generative model using \name.

% \textbf{Parallel development}: \name enables independent developments of language and vision modules, de-risking the complexities associated with a large-scale, joint multimodal pretraining. \lili{this one is very vague, first, it is not clear if we are comparing llamafusion vs transfusion scratch, or llamafusion vs dense finetuning, second, we didn't discuss which complexities. Consider move this to later discussion, and can expand on two points, 1) it is risky/unnecessary to scale image model to the same size as text model (400B). 2. dense finetuning normally has an adapter tuning stage, introducing complexity } 

% \section{Acknowledgement}

% future work: 
% 1. compose a pre-trained image encoder: (a) add a new CLIP encoder to Transfusion/LlamaFusion; (b) use pretrained multimodal LLMs that has incorporated vision encoder?
% 2. instruction tuning the model

% \han{Qwen model + Transfusion}

\bibliographystyle{assets/plainnat}
\bibliography{main}

\end{document}

%% TITLE
\title{\paperTitle}
\author{\authorBlock}
\maketitlesupplementary

\appendix
% \clearpage
% \section{Appendix Section}
% \label{sec:appendix_section}
% % Supplementary material goes here.

\clearpage
\setcounter{page}{1}
\maketitlesupplementary
\section{Additional Analysis}

\begin{figure*}[ht]
    \centering
    \includegraphics[width=\linewidth]{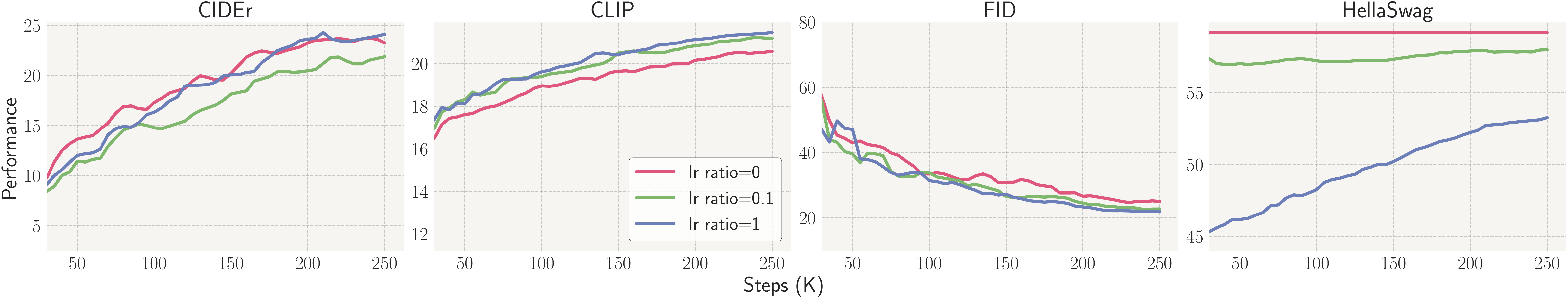}
    \caption{\textbf{Performance of shallow modality separation with varying lr ratios $\frac{\eta_{\text{text}}}{\eta_{\text{image}}}$.} }
        \label{fig:ablation_moe_performance}
\end{figure*}

\begin{figure*}[]
    \centering
    \includegraphics[width=\linewidth]{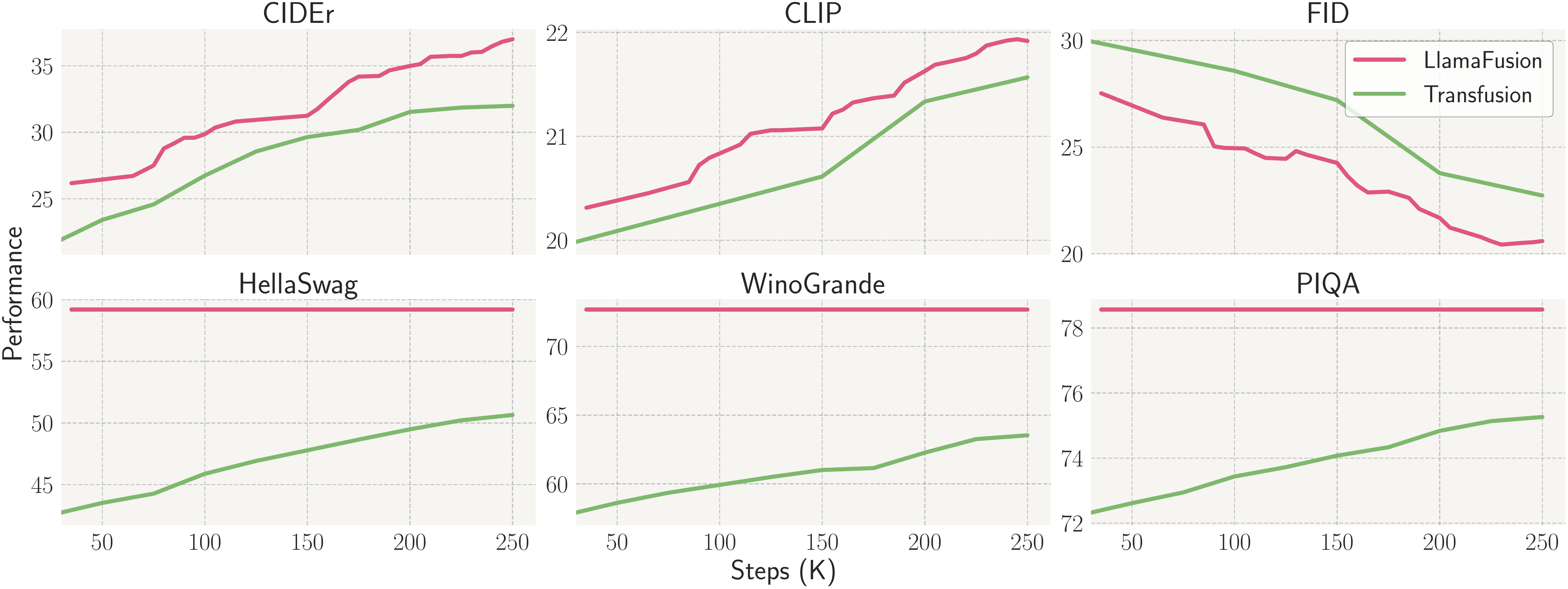}
    \caption{Evaluation of \name and Transfusion ($0.5 \times $ image FLOPs) during training.
    }
    \label{fig:performance_vs_token_bs1}
\end{figure*}

{\small
\bibliographystyle{ieeenat_fullname}
\bibliography{11_references}
}